\documentclass[runningheads]{llncs}
\usepackage[T1]{fontenc}
\usepackage{graphicx}
\usepackage{multirow}%
\usepackage{amsmath,amssymb,amsfonts}%
\usepackage{mathrsfs}%
\usepackage{xcolor}%
\usepackage{textcomp}%
\usepackage{manyfoot}%
\usepackage{booktabs}%
\usepackage{algorithm}%
\usepackage{algorithmicx}%
\usepackage{algpseudocode}%
\usepackage{listings}%
\usepackage{makecell}
\usepackage{array}
\usepackage{caption}
\usepackage{hyperref}

\begin{document}
\title{Echo State Networks for Bitcoin Time Series Prediction}
%
%
\author{Mansi Sharma\inst{1}\orcidID{0000-0001-9196-6342} \and
Enrico Sartor\inst{2}\orcidID{0009-0009-2697-6849} \and
Marc Cavazza\inst{3}\orcidID{0000-0001-6113-9696} \and
Helmut Prendinger\inst{4}\orcidID{0000-0003-4654-9835}}
\authorrunning{Sharma et al.}
%
\institute{DFKI, Saarland Informatics Campus, Germany \and
Politecnico di Milano, Italy \and University of Stirling, UK \and National Institute of Informatics, Japan \\
\email{mansi.sharma@dfki.de}}

%
\maketitle              
\abstract{
Forecasting stock and cryptocurrency prices is challenging due to high volatility and non-stationarity, influenced by factors like economic changes and market sentiment. Previous research shows that Echo State Networks (ESNs) can effectively model short-term stock market movements, capturing nonlinear patterns in dynamic data. To the best of our knowledge, this work is among the first to explore ESNs for cryptocurrency forecasting, especially during extreme volatility. We also conduct chaos analysis through the Lyapunov exponent in chaotic periods and show that our approach outperforms existing machine learning methods by a significant margin. Our findings are consistent with the Lyapunov exponent analysis, showing that ESNs are robust during chaotic periods and excel under high chaos compared to Boosting and  Na{\"i}ve methods.
}

\keywords{chaotic time series, cryptocurrency forecasting, echo state networks, lyapunov exponent}

%
%

\section{Introduction}\label{ch:chapter_one} 
\sloppy
Time series data are pervasive in both human activities and natural processes, including finance \cite{zhang,dan,lin}, energy consumption \cite{chou2018forecasting}, and climate variability \cite{ren2021deep}. Accurate forecasting supports decision-making, but volatility, non-linearity, and noise in real-world data challenge traditional models.
 \sloppy
Traditionally, forecasting used statistical models like ARIMA and Exponential Smoothing for their simplicity and interpretability \cite{bandara,smyl}, but these often fail to capture complex, non-linear patterns in large datasets. ML and DL methods offer adaptability by learning directly from data, though debates persist about their interpretability, efficiency, and scalability.

While DL architectures have demonstrated success in modeling complex temporal relationships their use in time series forecasting presents several challenges. These include high computational demands, longer training times, sensitivity to hyperparameter tuning, and a lack of interpretability. Moreover, many DL models require extensive training data to generalize well, making them less suitable for domains with limited or noisy time series data. Their black-box nature also hinders their adoption in critical applications where transparency is much needed.



In contrast, Echo State Networks (ESNs) offer a lightweight and computationally efficient alternative for time series (TS) forecasting \cite{practicalesn,zhang,jaeger}. 
As a type of Reservoir Computing model, ESNs leverage a fixed, randomly connected recurrent reservoir, where only the output weights are trained \cite{jaeger}. This approach significantly reduces training complexity and computational requirements. Because the internal dynamics of the reservoir are not adjusted during training, ESNs tend to be robust to overfitting and can generalize well, even in low-data or noisy settings \cite{practicalesn}. Their architecture naturally captures temporal dependencies without the need for backpropagation through time, making them efficient and scalable. 
Recent studies have shown the applicability of ESNs in various domains, including hydrological modeling, control systems, and financial forecasting \cite{li2015echo,lin,dan}. Despite their advantages, ESNs remain underexplored in the context of cryptocurrency forecasting—a domain characterized by extreme volatility, noise, and non-stationarity. Bitcoin, in particular, presents a highly challenging time series due to its stochastic behavior and susceptibility to abrupt market events.

As a use case, our work addresses the challenge of highly volatile Bitcoin prediction by forecasting its daily closing prices. The substantial price volatility and stochastic characteristics of Bitcoin time series data pose significant challenges in forecasting performance solely based on historical price movements. The need for more historical data specific to cryptocurrencies compounds these challenges, further complicating the identification of long-term patterns and trends over extended periods. 
Our specific contributions are threefold:

\begin{enumerate}
 \item {To the best of our knowledge, this is among the first explorations of ESN to forecast cryptocurrency prices. We show that with extensive fine-tuning of the ESN hyperparameters, it is possible to generate Bitcoin price predictions outperforming other mainstream ML methods.}
 \item {We evaluate ESNs using short time windows to stabilize the time series and compare their performance on uni-variate and multi-variate data, highlighting our comprehensive analysis of the choice of features for forecasting.}
  \item {We analyze results from the perspective of TS chaotic behavior, using Lyapunov’s exponent, suggesting a potential further criterion for ESN applicability.}  
\end{enumerate}

\section{Related Work} 
\label{sec:RW} 
Time series forecasting is widely used in fields such as finance, economics, weather prediction, supply chain management, and healthcare \cite{kenny,bandara,elsayed,fischer,jaquart}. Hybrid approaches combining statistical methods with ML algorithms have gained popularity for improving predictive accuracy \cite{smyl}. 
Makridakis \emph{et al.} \cite{makridakis18} argued that ML methods can identify data relationships automatically, unlike statistical methods that require trend and seasonality specifications. However, ML performance depends on data availability and often lags behind traditional statistical methods in non-stationary time series \cite{spiliotis}. 
Makridakis \emph{et al.} found that statistical methods generally outperformed ML across accuracy measures and forecasting horizons. Hybrid models combining ML and statistical techniques have shown improved forecasting accuracy \cite{smyl}. Makridakis \emph{et al.} \cite{makridakis22} later found that ensemble deep learning models perform better for long-range predictions, while statistical and ML models excel in short-term forecasting.
Previous research mainly focuses on price trend classification and time series regression. However, evaluation methods vary, leading to conflicting results. We provide a comprehensive analysis of cryptocurrency forecasting across both tasks.

\subsection{Cryptocurrency Forecasting}
There is growing interest in cryptocurrencies, particularly with Bitcoin leading the way as the first digital currency \cite{nakamoto}. 
\sloppy
Studies have assessed Bitcoin’s market efficiency over time with mixed findings. Urquhart \cite{urquhart} found inefficiency from 2010–2016, while Nadarajah \cite{nadarajah} observed weak form efficiency using power transformations. Likewise, Bariviera \cite{bariviera} showed Bitcoin market was not weak form efficient before 2014, but became weak form efficient afterwards. Overall, evidence is mixed, though many suggest Bitcoin has become more efficient as the market grows.

\textbf{Classification}. 
Borges \emph{et al.} \cite{borges} classify trends by resampling data using closing price thresholds and averaging results from four models: Logistic Regression and Support Vector Machine (linear), and Random Forest and Decision Tree Gradient Boosting (non-linear). While ensemble methods often improve regression problems, ESNs offer computational efficiency and simplicity, requiring only the output layer to be trained. ESNs excel at capturing temporal dependencies and non-linear patterns, making them ideal for volatile markets like cryptocurrency. Sun \emph{et al.} \cite{sun} show that LightGBM outperforms classical ML methods on a dataset of 42 cryptocurrencies. Jaquart \emph{et al.} \cite{jaquart} predict trends on a high-frequency dataset using various ML and DL models, finding that RNNs and Gradient Boosting provide the best results.
\\
\textbf{Regression}. 
Alessandretti \emph{et al.} \cite{alessandretti} compare two XGBoost models and an LSTM to a baseline Simple Moving Averages model, showing that LSTMs excel in long-range predictions while XGBoost performs well for short-term predictions with a small window of recent data. Elsayed \emph{et al.} \cite{elsayed} enhance XGBoost predictions by adding previous data points as features, an approach we adapted for our model. Wood \emph{et al.} \cite{wood} developed a custom transformer model for time series forecasting, but its extended training time is a downside. Jaeger \emph{et al.} \cite{jaeger} first applied ESNs to stock prediction in 2009, using the Hurst exponent and PCA to improve performance. Lin \emph{et al.} \cite{lin} extended this to predict stock market returns, demonstrating ESNs' outperformance of other neural networks, though PCA was necessary for some stocks. Dan \emph{et al.} \cite{dan} tested a deterministic ESN on the Chinese stock market and S\&P 500, showing modest improvements. Zhang \emph{et al.} \cite{zhang} demonstrated the competitiveness of ESNs compared to other ML and DL methods.
Kenny \cite{kenny} is the only study to apply ESNs to cryptocurrency forecasting but without extensive ESN tuning, which is the focus of our work. 

\begin{figure}[t]
    \centering
   \includegraphics[trim={0.5cm 0cm 3cm 1cm},clip,width=0.6\linewidth]{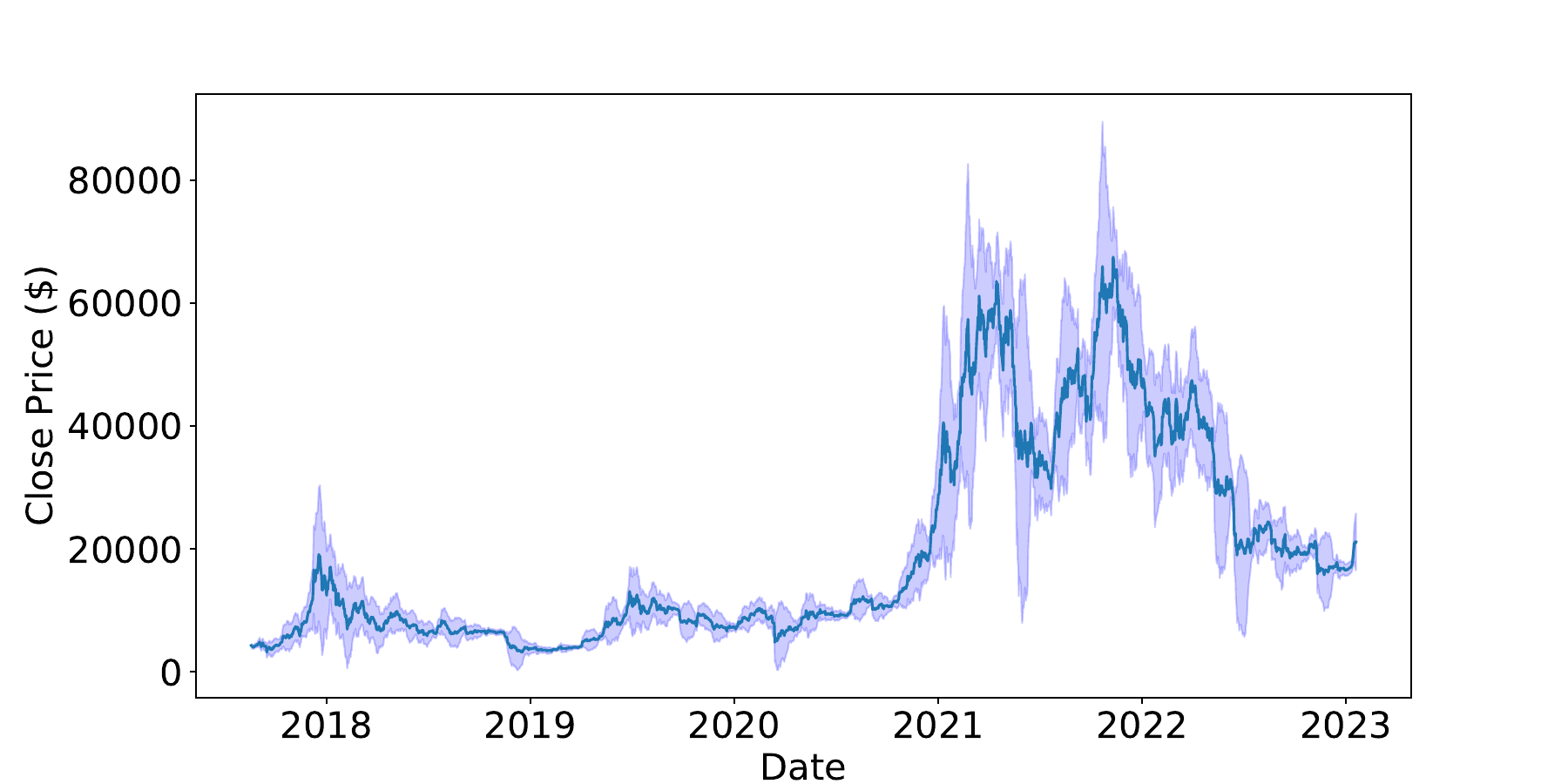} 
   \captionsetup{width=0.8\linewidth}
  \caption{Bitcoin data showing the closing prices with 
 standard deviation computed on a moving windows of 30 days.}
    \label{fig:btcstd}
\end{figure}

\section{Dataset and Descriptive Statistics} 
\label{sec:sec22}
We used daily Bitcoin data evaluated in US dol~lars (ticker: BTCUSDT) sourced from the Binance trading platform \cite{binance}. Missing values were filled using linear interpolation, with no additional transformations applied, though we considered standardization and converting prices to returns or log returns. Our preliminary experiments indicated that returns and log returns, despite their stationarity, present higher variance and are more challenging to predict. Standardization or scaling could facilitate model comparison, but we chose to avoid these preprocessing methods to maintain clarity regarding our models' performance, as even basic standardization could significantly affect the dataset due to non-stationarity. 

The dataset spans from August 17\textsuperscript{th}, 2017 to January 24\textsuperscript{th}, 2023, and includes opening, closing, high, and low prices. The opening price is recorded at the start of the day, while the closing price is recorded at the end. Figure \ref{fig:btcstd} illustrates the closing prices and their standard deviation, 
\begin{figure}[t]
    \centering 
   \includegraphics[trim={0cm 0cm 0cm 0cm},clip,width=0.6\linewidth]{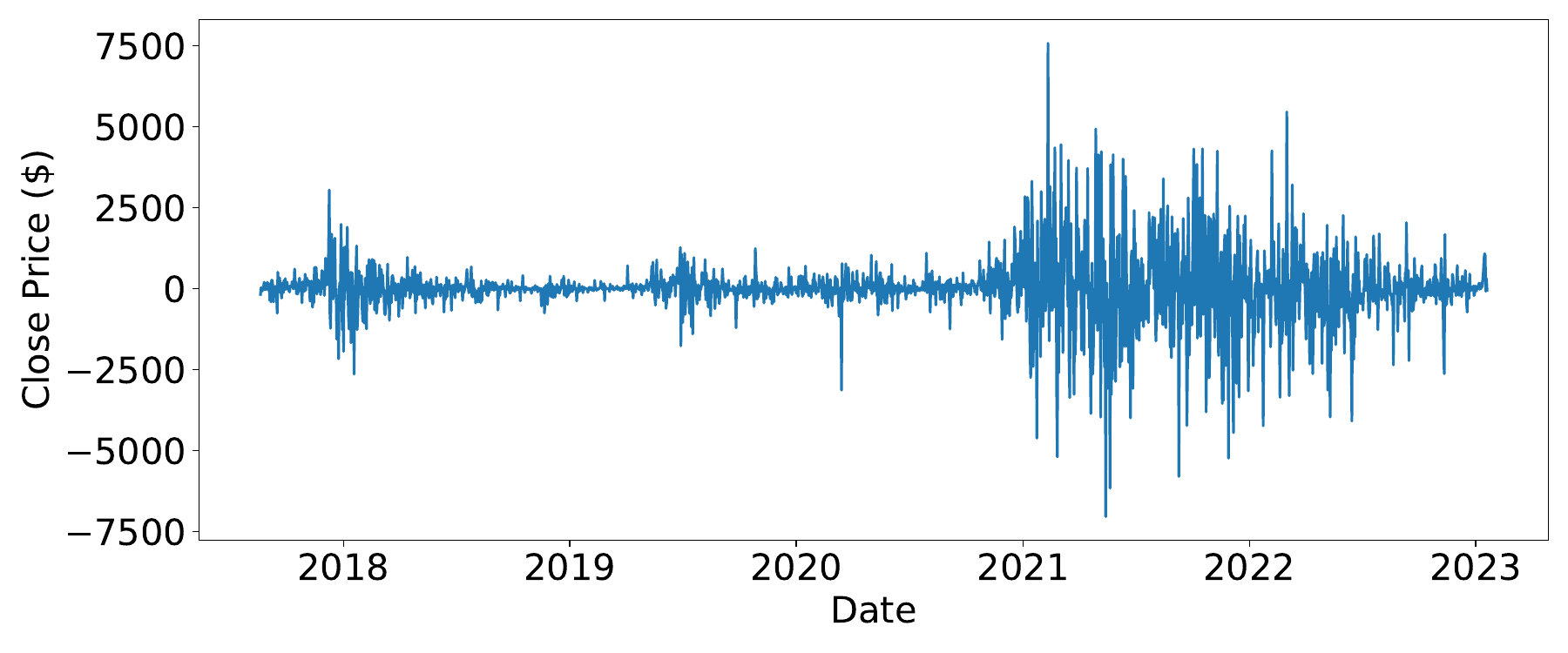} 
    \caption{Bitcoin differenced prices.}
    \label{fig:diff}
\end{figure}
computed using a 30-day moving window, represented as a band with a thickness of 3 standard deviations on each side. We opted for daily data because the results were almost independent of frequency, a phenomenon aligned with the Fractal Market Hypothesis (FMH) \cite{fract}, which suggests that market behavior exhibits consistent patterns regardless of the time scale. The forecasting task focuses on predicting the closing price for the following day, with an emphasis on one-day-ahead forecasts.

Figure~\ref{fig:btcstd} shows various types of non-stationarities in the data. These include linear and exponential trends (with the latter emerging towards the end of 2020), abrupt changes indicative of structural breaks (such as at the start of 2021), heteroscedasticity, characterized by fluctuating variance over time, and a non-normal, non-symmetric price distribution with fat tails.

To address these issues, we performed fractional differentiation, similar to the approach used by \cite{lopez}, to determine the minimum level of differentiation required to achieve stationarity according to the Augmented Dickey-Fuller (ADF) test, while preserving as much memory as possible. The analysis revealed a fractional differentiation order of 0.8, indicating that the integration order is less than 1. Despite this, the data remains non-stationary at higher moments (as depicted in Figure~\ref{fig:diff}), indicating the complexity of the forecasting challenge. Cryptocurrency markets, particularly Bitcoin, are known for their high volatility. This volatility stems from a combination of factors, including speculative trading, regulatory news, market sentiment, and the relatively immature state of the market compared to traditional assets like stocks and bonds. Such extreme price fluctuations pose a significant challenge for forecasting models, as they need to capture the underlying chaotic behavior of the market. Our study aims to leverage advanced forecasting techniques, specifically ESNs, which are well-suited for handling non-linear and chaotic time series data.  


\section{Methods}\label{sec:methods} 

In this section, we describe the Echo State Network approach, detail the chaos analysis using the Lyapunov exponent, and explain our feature selection process. Additionally, we present the baseline for comparison.

\subsection{Echo State Network} \label{ESN}
Echo State Networks (ESN) belong to the category of recurrent neural networks and are inherently chaotic due to the random connections between their neurons. ESN was originally proposed by Jager et al.~\cite{jaeger} to tackle nonlinear system learning and predict chaotic time series. 

\begin{figure}[t]
    \centering
   \includegraphics[trim={0cm 0cm 0cm 0cm},clip,width=0.6\textwidth]{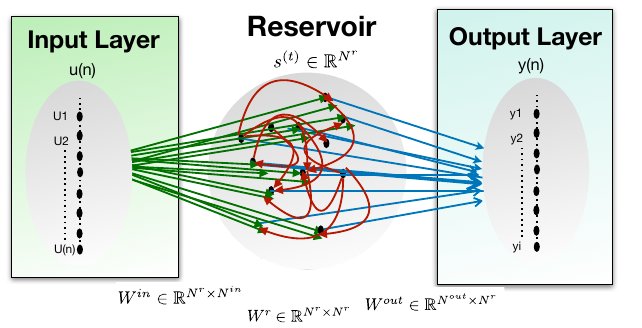} 
   \captionsetup{width=0.8\linewidth}
    \caption{Building block of ESN: layered structure, core parameters, and their dimensionality. Figure taken from Li \emph{et al.} \cite{li2015echo}}
    \label{fig:ESN_archi}
\end{figure}

During training, Echo State Networks (ESNs) update only the connections between the reservoir (observer) and the output. The internal network forms a large reservoir of coupled oscillators, where input drives their states, and predictions rely on their responses. A key advantage of ESNs is that only the output connections are trained—hidden and input connections remain fixed. Figure~\ref{fig:ESN_archi} illustrates the architecture of the ESN model, which is described as follows:
\begin{align}\label{eq:esn1}
\tilde{x}(k) = \tanh (W_{in} u(k) + W_x x(k-1))
\end{align}
\begin{align}\label{eq:esn2}
y(k) = W_{out} (\alpha \tilde{x}(k) + (1-\alpha) x(k-1)) 
\end{align}
where $\alpha$ is the leaking rate, a hyper-parameter. The two equations above represent, respectively, the dynamic and output equations of the ESN, where $x(k)$ stands for the internal state vector of the reservoir, $u(k)$ and $y(k)$ are the input vector and model output, respectively. The hyperbolic tangent activation function is denoted by $\tanh$, $W_x$ represents the internal connection weight matrix of the reservoir, $W_{in}$ denotes the input weight matrix, and $W_{out}$ is the output weight matrix. 

The only trainable part of the ESN is the output weight matrix $W_{out}$, which can be determined using a simple linear regression as shown in:
\begin{align} \label{eq:phiw}
y = \Phi W_{out}+\epsilon y 
\end{align}
where
\begin{align} \label{eq:phiw1}
\Phi = [x(k), x(k+1), ..,x(k+N-1)]^T
\end{align}
\begin{align} \label{eq:phiw2}
y = [y(k), y(k+1), .., y(k+N-1)]^T
\end{align}
In Equations \eqref{eq:phiw1} and \eqref{eq:phiw2}, $k$ is the starting index of the training samples, typically set to discard the influence of the reservoir's initial transient. $\epsilon$ is assumed to be a zero-mean Gaussian noise with variance $\beta$, and $N$ is the number of training samples. We implement ESN using \textit{reservoirpy} \footnote{https://reservoirpy.readthedocs.io/en/latest/} and tune the hyper-parameters values as inspired by \cite{practicalesn}.

\subsection{Chaos Analysis}
We compute the Maximal Lyapunov Exponent (MLE) using Rosenstein et al.'s algorithm \cite{rosenstein} to quantify chaos in a time series. A positive MLE indicates strong chaos, zero indicates a limit cycle or quasi-periodic orbit, and a negative MLE suggests a fixed point. The Lyapunov exponent is described as follows.
For an m-dimensional system, there are $m$ Lyapunov exponents denoted as $\lambda_1, \lambda_2, ..., \lambda_m$ in descending order of significance.
\begin{table}[t]
\captionsetup{width=0.8\linewidth}
\caption{Tuned ESN parameters. Dist. refers to the type of distribution. Step applies only when relevant.}
\label{table:ESN}
\centering
\renewcommand{\arraystretch}{1}
\setlength{\tabcolsep}{2pt}
\begin{tabular}{|c|c|c|c|c|}
\hline
\textbf{Parameter(s)} & \textbf{Dist.} & \textbf{From} & \textbf{To} & \textbf{Step} \\ \hline
Reservoir Size & \makecell{Discrete\\Uniform} & 50 & 250 & 50 \\ \hline
\makecell[l]{Spectral Radius, Leaking Rate, \\ Output Weight Reg., Reservoir Noise, \\ Input Noise, Feedback Noise} & \makecell{Log\\Uniform} & \makecell{$10^{-7}$/\\$10^{-6}$/$10^{-4}$} & \makecell{0.1/\\1/10} & -- \\ \hline
\makecell[l]{Input Scaling, Bias Scaling, Feedback \\ Scaling, Sparsity} & Uniform & 0.05/0.8 & 0.5/1 & 0.1 \\ \hline
\makecell[l]{Noise Type (Gaussian), Seed (2023)} & Fixed & -- & -- & -- \\ \hline
\end{tabular}
\end{table}
Consider a dynamical system described by:
\begin{align} \label{eq:1.2}
    \frac{d\textbf{x}(t)}{dt} = \textbf{F}(\textbf{x}(t))
\end{align}
Taking variation of both sides of (\eqref{eq:1.2}) yields the following equation governing the evolution of the infinitesimal vector $\delta \textbf{x}$ in the tangent space at $\textbf{x}(t)$:
\begin{align} \label{eq:2.7} 
    \frac{d\delta\textbf{x}}{dt} = \frac{\partial \textbf{F}}{\partial \textbf{x}} \cdot \delta \textbf{x}
\end{align}
Solving for (\eqref{eq:2.7}) gives:
\begin{align}
    \delta \textbf{x}(t) = \textbf{A}^t \delta \textbf{x}(0)
\end{align}
where $\textbf{A}^t = exp( \int \frac{\partial \textbf{F}}{\partial \textbf{x}}	dt)$ is a linear operator that evolves an infinitesimal vector from time $0$ to time $t$. The mean exponential rate of divergence of the tangent vector is then given by:
\begin{align}
    \lambda[\textbf{x}(0),\delta \textbf{x}(0)] = \lim_{t\to\infty} \frac{1}{t} ln |\frac{\delta \textbf{x}(t)}{\delta \textbf{x}(0)}|
\end{align}
The Lyapunov exponents are given as:
\begin{align}
    \lambda_i = \lambda[\textbf{x}(0), \textbf{e}_i]
\end{align}
where $e_i$ is an m-dimensional basis vector.

Each Lyapunov exponent, denoted as $\lambda_i$, represents the average divergence rate over the entire attractor and is computed based on the m-dimensional basis vector $e_i$. The values of $\lambda_i$ for chaotic systems are independent of the initial condition $\textbf{x}(0)$ due to the property of ergodicity. To determine whether a time series exhibits chaotic behavior, one primarily focuses on calculating the maximal Lyapunov exponent, $\lambda_1$, which is denoted as $\lambda_{max}$. 

\subsection{Features} \label{sec:features}
Jaquart \emph{et al.}~\cite{jaquart2} categorized crypto features into four groups: technical, blockchain-based, sentiment/interest-based, and asset-based. Technical features, such as Bitcoin returns and indicators, were prioritized to improve forecasting accuracy by focusing on price-related data and minimizing external noise. We selected technical features based on market knowledge, expert input, and experimental analysis, rather than automated methods. Features were evaluated for their impact on prediction accuracy across time windows, ensuring computational efficiency while maintaining relevance. 
We use the 3- and 7-day moving averages of the closing price, the 7-day closing price standard deviation, the difference between the high and low prices, and the difference between the closing and opening prices. Additionally, we compute the ratio of the last closing price to the 7-day moving average, the trend coefficients of the regression line over the last 7 and 14 closing prices, the 7-day Relative Strength Index (RSI), and the 7-21 period Moving Average Convergence/Divergence.

\section{Evaluation} 
\label{ch:chapter_three} 
For a highly non-stationary dataset, we use a rolling-window cross-validation approach. The training set size remains fixed while the window slides forward, dropping the oldest data for each new forecast. This ensures that each model is evaluated on the same number of training points. We pre-split the data into windows with training sizes of 15, 30, and 60 points, a test size of 10 points, and a stride of 1.

\textbf{Evaluation Metrics}
We used the classical Root Mean Squared Error (RMSE) (\eqref{eq:rmse}). Further, it is easy to understand and popular in the previous studies \cite{kenny,lin}. 
\begin{align} \label{eq:rmse}
RMSE = \sqrt{\frac{1}{n}\sum_{t=1}^n(y_t-\hat{y_t})^2}
\end{align}
Note that the RMSE is always computed over a test set of length 10 days (10 points) and is measured in US dollars.

\textbf{Hyperparameter Selection} Both ESN and XGBoost require tuning of their respective hyperparameters to achieve optimal results. We employed Bayesian Optimization, specifically Tree Parzen Estimators (TPE), using the widely adopted $hyperopt$ \footnote{https://hyperopt.github.io/hyperopt/} Python package, to search for the best hyperparameter combinations in their spaces. Table~\ref{table:ESN} and Table~\ref{table:XGB} describe the list of tuned parameters for ESN and XGB, respectively.

\begin{table}
\captionsetup{width=0.8\linewidth}
\caption{Tuned XGB parameters. Dist. refers to the sampling distribution.}
\label{table:XGB}
\centering
\renewcommand{\arraystretch}{1}
\setlength{\tabcolsep}{5pt}
\begin{tabular}{|l|c|c|c|c|}
\hline
\textbf{Hyper-parameters} & \textbf{Dist} & \textbf{From} & \textbf{To} & \textbf{Step} \\ \hline
\multicolumn{5}{|c|}{\textbf{Discrete Uniform}} \\ \hline
No. of Estimators &  & 100 & 300 & 100 \\
Max Depth &  & 3 & 15 & 1 \\
Gamma &  & $10^{-4}$ & 10 & $10^{-3}$ \\
Learning Rate &  & $10^{-2}$ & 0.3 & $10^{-2}$ \\
Col. Sample by Tree &  & 0.5 & 1 & 0.1 \\
Min Child Weight &  & 0 & 10 & 1 \\
Subsample &  & 0.8 & 1 & 0.1 \\ \hline
\multicolumn{5}{|c|}{\textbf{Fixed}} \\ \hline
Random State & Fixed & 2023 &  &  \\ \hline
\end{tabular}
\end{table}

\subsection{Baseline}
The classic baseline in time series forecasting is known as the \textbf{Last Value} or simply the \textbf{Na{\"i}ve Method}. For one-step ahead forecasts, this approach uses the current price as the prediction for the next value. 
Extreme Gradient Boosting (XGBoost) is widely used for time series forecasting \cite{alessandretti,kenny}. Building on Gradient Boosting Decision Trees (GBDT), it improves the cost function using second-order Taylor expansion for greater precision.
\begin{figure*}[t]
    \centering
   \includegraphics[trim={0cm 0cm 0cm 0cm},clip,width=\linewidth]{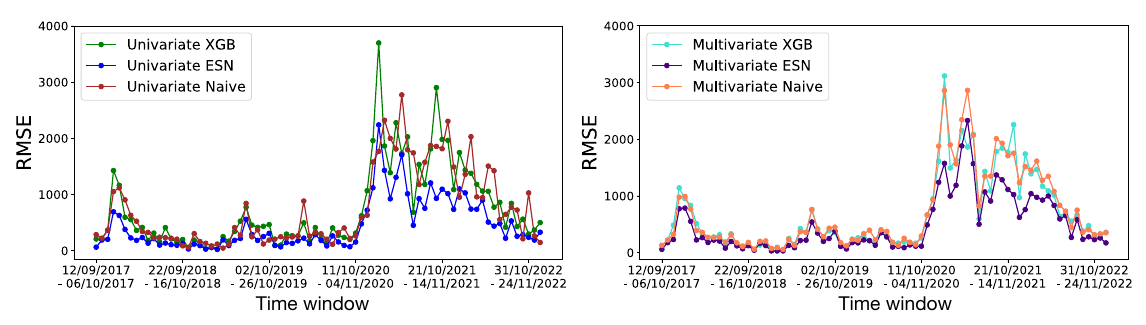} 
   \captionsetup{width=0.8\linewidth}
    \caption{RMSE values on uni- and multi-variate setting using a 15-day training window across various time periods.} 
    \label{fig:uni_rmse15}
\end{figure*}
In addition to the baseline methods, we experimented with other traditional statistical models such as ARIMA, Exponential Smoothing, and regression-based approaches, like Generalized Linear Models, Huber Regression, and Weighted Least Squares. However, their RMSE ranged from $800$ to $2000$, performing significantly worse than other methods. Thus, we focused on reporting the two best-performing baselines.

\begin{figure*}
    \centering
   \includegraphics[trim={0cm 0cm 0cm 0cm},clip,width=\linewidth]{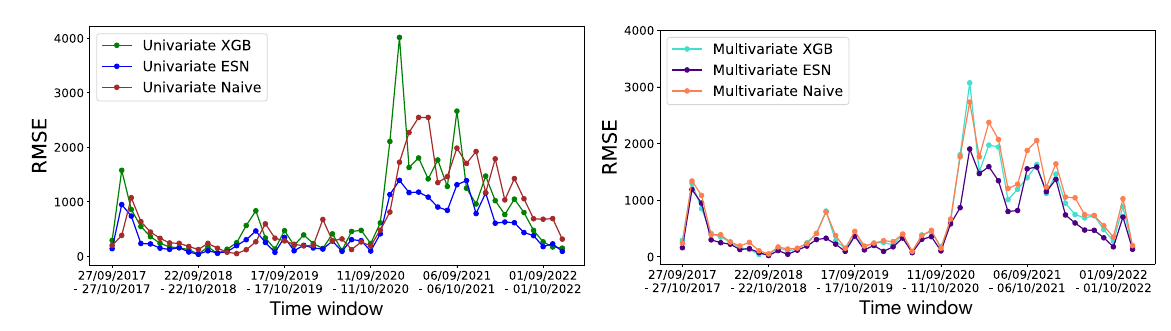} 
   \captionsetup{width=0.8\linewidth}
    \caption{RMSE values on uni- and multi-variate setting using a 30-day training window across various time periods.} 
    \label{fig:uni_rmse30}
\end{figure*}

\begin{figure*}
    \centering
   \includegraphics[trim={0cm 0cm 0cm 0cm},clip,width=\linewidth]{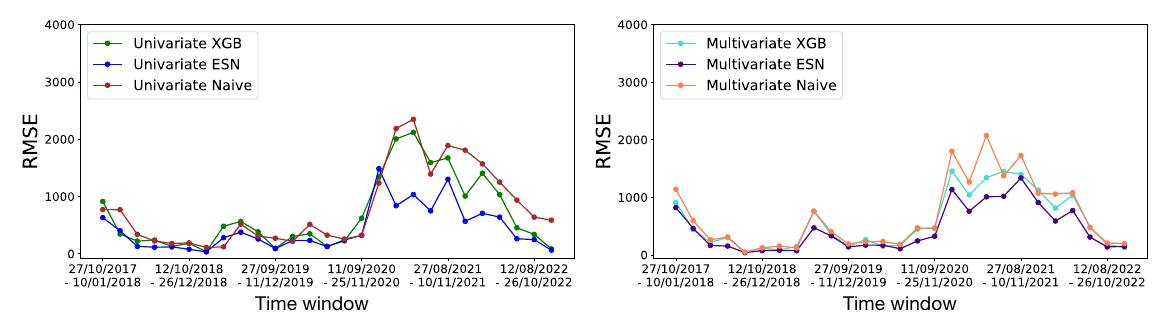} 
   \captionsetup{width=0.8\linewidth}
    \caption{RMSE values on uni- and multi-variate setting using a 60-day training window across various time periods.} 
    \label{fig:uni_rmse60}
\end{figure*}
\section{Results} 
\label{ch:results}



We present results on the uni-variate data consisting of the closing prices and on a one-step ahead prediction task and also on multi-variate data with the technical features described in Section~\ref{sec:features}. To conclude, we present an analysis of the Lyapunov exponent and show that the ESN performs significantly better than XGBoost on the most chaotic windows. 


\textbf{Uni-variate data} We compare ESN with baseline approaches, XGBoost and Last Value on uni-variate data (Bitcoin closing prices) and on windows of training size 15, 30, 60 days. 

Figure~\ref{fig:uni_rmse15} (left), Figure~\ref{fig:uni_rmse30} (left), and Figure~\ref{fig:uni_rmse60} (left) shows the RMSEs for the years 2017--2022 for all three training windows. Table~\ref{tab:rmse_values} shows mean RMSE values for different training window sizes and uni-variate setting, highlighting that ESN achieves lowest RMSE values among all the windows, especially with larger window sizes, surpassing other methods. 

Note that the RMSE values seem high in absolute value, but they are quite low in percentage, as in the year 2022 Bitcoin had high price values, compared to the rest of its history. In fact, in percentage with respect to the closing price, the ESNs' RMSEs range between 1.0\% and 2.7\%, with an average value of 1.8\%. The XGBoost model, although competitive, shows significant peaks in RMSE, particularly around late 2020 and early 2021, suggesting that it struggles more during this challenging period. The Na{\"i}ve Model, serving as a baseline, consistently shows higher RMSE values, reflecting its relative simplicity and lower predictive power.

\begin{figure*}
    \centering
   \includegraphics[trim={0cm 0cm 0cm 0cm},clip,width=0.8\linewidth]{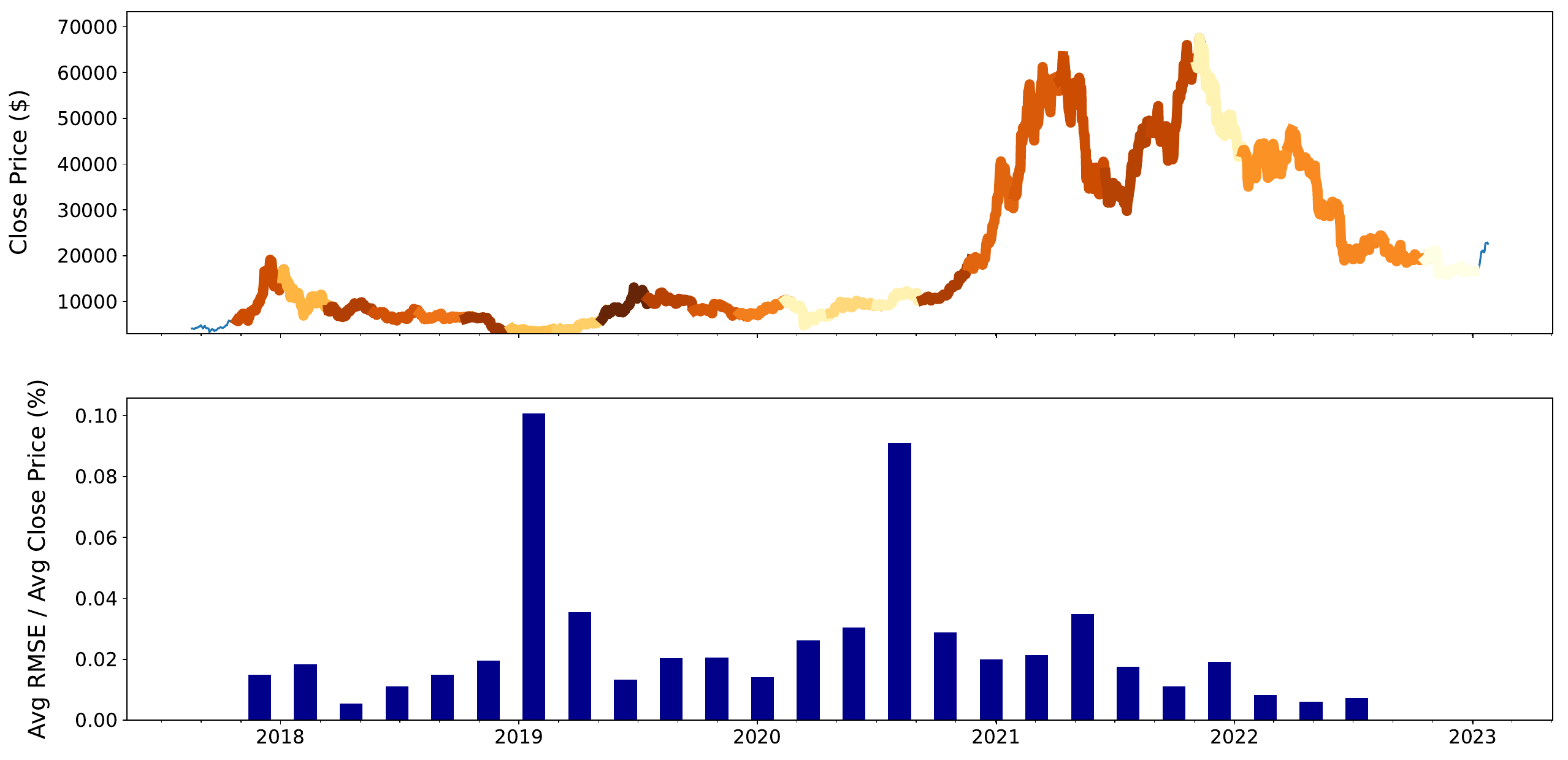} 
   \captionsetup{width=0.8\linewidth}
    \caption{(Top) Amount of chaos measured by Lyapunov exponent, year by year from 2018 to 2022 using window size 60. The darker the color, the higher the chaos. (Bottom) Ratio of ESN's RMSE and closing price year by year using window size 60.} 
    \label{fig:lyp}
\end{figure*}

\textbf{Multi-variate data} We compare ESN with baseline approaches, XGBoost and Last Value, on multi-variate data on windows of training size 15, 30, 60 days. The data consists of the Bitcoin closing prices plus the features introduced in Section~\ref{sec:features}. 

Table~\ref{tab:rmse_values} presents the RMSE values for various training windows on multivariate data. Similar to the results from uni-variate data predictions,  ESN consistently achieves the lowest RMSE value, demonstrating superior performance. Figure~\ref{fig:uni_rmse15} (Right), Figure~\ref{fig:uni_rmse30} (Right), and Figure~\ref{fig:uni_rmse60} (Right) shows the RMSEs for the years 2017--2022 for all three training windows. The multi-variate models generally perform better than their uni-variate counterparts, as evidenced by the lower RMSE values in the bottom row of each set. This is particularly true for the ESN and XGBoost models, where the inclusion of multiple variables seems to enhance the accuracy of the predictions. However, the Na{\"i}ve Model's performance remains comparatively poor even in the multi-variate setting.

\begin{table}[h]
\centering
\captionsetup{width=0.8\linewidth}
\caption{Mean RMSE for uni- and multi-variate settings across different window sizes.}
\label{tab:rmse_values}
\begin{tabular}{|l|ccc|ccc|}
\hline
\multirow{2}{*}{\textbf{Model}} & \multicolumn{3}{c|}{\textbf{Uni-variate}} & \multicolumn{3}{c|}{\textbf{Multi-variate}} \\ \cline{2-7}
                                & 15     & 30     & 60     & 15     & 30     & 60     \\ \hline
XGB                             & 740.43 & 741.84 & 675.95 & 678.59 & 676.60 & 572.58 \\
ESN                             & \textbf{442.78} & \textbf{463.96} & \textbf{427.35} & \textbf{487.77} & \textbf{533.43} & \textbf{441.23} \\
Na{\"i}ve                       & 736.60 & 751.18 & 787.02 & 713.53 & 749.96 & 665.62 \\ \hline
\end{tabular}
\end{table}

\subsection{Chaos analysis} 
\label{sub:comp4} 
Figure~\ref{fig:lyp} illustrates the challenges of predicting Bitcoin's closing prices using Echo State Networks, particularly during chaotic periods characterized by high Lyapunov exponents. Figure~\ref{fig:lyp} (Top)  suggests periods of different dynamics or market behavior, potentially indicating chaotic behavior in certain periods year by year. Darker colors indicate higher levels of chaos. Figure~\ref{fig:lyp} (Bottom)  shows the ratio of ESN's RMSE and closing price year by year using window size 60. The most prominent peaks appear around early 2019 and early 2021, indicating that the Bitcoin market was highly volatile and unpredictable during these times, likely reflecting chaotic behavior as suggested by a high Lyapunov exponent. 

We present the statistical tests on whether ESN works significantly better than XGBoost in the windows where there is higher chaos, as measured by Lyapunov exponent. We split the dataset into windows of training size 15, 30, 60 days and run the following experiment for each of the three scenarios. We compute, for each window, the maximal Lyapunov exponent and the ratio between the RMSE obtained with ESN and the RMSE obtained with XGBoost, we call this EXratio. 

\begin{enumerate}
    \item When the EXratio is much smaller than 1, it means that the ESN performs much better than XGBoost;
    \item When the EXratio value approaches 1, it performs just slightly better than XGBoost.
\end{enumerate}


Since we are using a cross-validation inside each window, the Lyapunov exponent is computed 10 times, one for each train-test split, and then averaged. 
To run the statistical test, we split the windows into two groups, one containing the most chaotic windows and one with the least chaotic. We take the EXratio values in each of the two groups and we test if their median is different in the two groups. 

\begin{figure}
    \centering
    \includegraphics[trim = {0cm 0cm 0cm 0cm}, clip, width=0.8\linewidth]{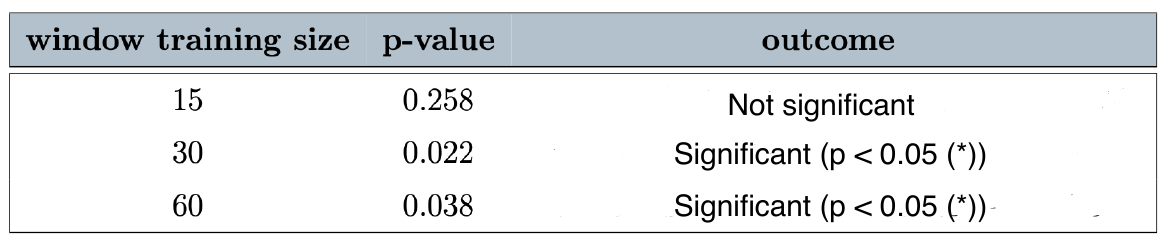}
    \caption{Chaos analysis}
    \label{table:mannU}
\end{figure}

To sum up, in statistical terms:

\begin{align*}
    H_0: \text{median(EXratio)}_{\textit{lc}} &= \text{median(EXratio)}_{\textit{hc}} \\
    H_1: \text{median(EXratio)}_{\textit{lc}} &\neq \text{median(EXratio)}_{\textit{hc}}
\end{align*}

where \textit{lc} refers to low chaos and \textit{hc} refers to high chaos. We tested the gaussianity of the samples to verify the assumption of the t-test, but they are not verified, so we use a Mann-Whitney U test with a significance of $5\%$.
Figure~\ref{table:mannU} presents results for training windows of sizes 30 and 60, confirming the hypothesis, whereas this is not observed for window size 15. This discrepancy might be due to the short duration of these windows, posing challenges in accurately computing the Lyapunov exponent with limited points available for constructing trajectories in the embedding space.



\section{Discussion}

We showed that ESN is particularly effective in forecasting Bitcoin prices during highly chaotic market conditions, as highlighted by the Lyapunov exponent analysis. This ability to handle non-linear and chaotic time series data, which are prevalent in financial markets, makes ESNs highly suitable for financial forecasting tasks, particularly in environments characterized by high volatility and unpredictability. Our comparative analysis demonstrated that ESNs consistently outperformed other state-of-the-art methods, such as XGBoost, especially in chaotic conditions. This superior performance indicates that ESNs are better at capturing the complex, underlying patterns in data than traditional machine learning approaches, which often struggle with the non-stationarity and noise inherent in financial time series. 


ESNs excel in capturing complex, non-linear temporal patterns, making them well-suited for volatile financial markets where traditional models often struggle. In real-world trading strategies, ESNs can be used to predict asset prices, detect trends, and even optimize trading strategies in real-time. For example, they could help traders anticipate short-term price movements in stocks, commodities, or cryptocurrencies, enabling more informed decision-making. The ability of ESNs to handle large volumes of data and adapt to changing market conditions without extensive retraining gives them a significant advantage in fast-paced environments. 


\section{Limitations} 
In our paper, we introduced the use of Echo State Networks  for forecasting Bitcoin prices, addressing the challenges posed by the market's volatility. As this is one of the first studies of its kind, it inherently has some limitations that should be explored in future research. One limitation is the use of relatively short time windows for stabilizing the time series data. While this approach helps to manage the short-term volatility of Bitcoin prices, it may not fully capture long-term trends or seasonal patterns that could enhance prediction accuracy. 


Another area to be explored is the multi-step ahead forecasting, which would involve predicting Bitcoin prices over multiple time horizons rather than just the next value. Multi-step forecasting is a more challenging task as it requires the model to maintain accuracy over a longer sequence of predictions, and errors can accumulate across multiple steps. Extending the ESN approach to multi-step ahead forecasting could offer deeper insights into Bitcoin price trends and provide more practical predictive capabilities, particularly for applications requiring long-term forecasts. 

\section{Conclusions and Future Work} 
\label{ch:conclusions}

This study explores the use of Reservoir Computing, particularly ESNs for forecasting daily cryptocurrency closing prices. ESNs proved effective across various market conditions including trends, sideways movements, and level shifts—by using a window-splitting approach that addressed volatility and non-stationarity. The results show that ESNs perform especially well in chaotic markets, as measured by the Lyapunov exponent, challenging the Efficient Markets Hypothesis (EMH) by demonstrating predictability in volatile financial series.

This work contributes empirical evidence for the viability of ESNs in cryptocurrency forecasting and opens new research directions. Future efforts may extend this approach to other cryptocurrencies, integrate lagged external features (e.g., via Convergent Cross Mapping), and explore techniques like automatic feature selection, multi-step forecasting, or trend prediction. While not designed as a trading algorithm, the study highlights the strong potential of ESNs under complex market dynamics.

\bibliographystyle{splncs04}
\bibliography{ref}
%




\end{document}